\documentclass{article}
\usepackage{ijcai25}

\usepackage{times}
\usepackage{soul}
\usepackage{url}
\usepackage[hidelinks]{hyperref}
\usepackage[utf8]{inputenc}
\usepackage[small]{caption}
\usepackage{graphicx}
\usepackage{amsmath}
\usepackage{amsthm}
\usepackage{booktabs}
\usepackage{algorithm}
\usepackage{algorithmic}
\usepackage[switch]{lineno}

\usepackage{hyperref}
\usepackage{url}

\usepackage{amsmath,amssymb,amsfonts}
\usepackage{textcomp}

\usepackage{booktabs}
\usepackage{algorithm}
\usepackage{mathtools}
\usepackage{array}
\usepackage{multirow}
\usepackage{stmaryrd}

\usepackage{wrapfig}
\usepackage{enumitem}

\usepackage{multirow}
\usepackage[table,xcdraw]{xcolor}

\usepackage{amsmath}
\usepackage{pifont}

\DeclareMathOperator*{\argmin}{arg\,min}

\newcommand{\xmark}{\ding{55}}%

\title{Latest Advancements Towards Catastrophic Forgetting under Data Scarcity: A Comprehensive Survey on Few-Shot Class Incremental Learning}

\author{
M. Anwar Ma'sum$^1$
\and
Mahardhika Pratama$^1$\and
Igor Skrjanc$^{2}$\\
\affiliations
$^1$University of South Australia, 
$^2$University of Ljubljana\\
\emails
masmy039@mymail.unisa.edu.au, dhika.pratama@unisa.edu.au,
igor.skrjanc@fe.uni-lj.si
}

\begin{document}

\maketitle
\begin{abstract}
Data scarcity significantly complicates the continual learning problem, i.e., how a deep neural network learns in dynamic environments with very few samples. However, the latest progress of few-shot class incremental learning (FSCIL) methods and related studies show insightful knowledge on how to tackle the problem. This paper presents a comprehensive survey on FSCIL that highlights several important aspects i.e. comprehensive and formal objectives of FSCIL approaches, the importance of prototype rectifications, the new learning paradigms based on pre-trained model and language-guided mechanism, the deeper analysis of FSCIL performance metrics and evaluation, and the practical contexts of FSCIL in various areas. Our extensive discussion presents the open challenges, potential solutions, and future directions of FSCIL. 

\end{abstract}

\section{Introduction}
The few-shot class incremental learning (FSCIL) problem simulates a real-world problem where a machine learning model is deployed to learn dynamic environments with very few samples \cite{TOPIC_tao2020few}. FSCIL problem is broadly applied in various area e.g. medical\cite{MEDMAPIC_10050007}, remote sensing\cite{GRSCUC_zhao2022few}, audio processing\cite{NLPAUDIODFSL_wang2021few}, and data mining \cite{GRAPH_tan2022graph}.  The presence of the data scarcity constraint in FSCIL hardens the catastrophic forgetting problem to be solved. As a clear proof, rehearsal-based class incremental learning (CIL) methods \cite{XDER_boschini2022class}, and event joint-training mechanism fail to overcome the FSCIL problem. Thus, it calls for advanced solutions beyond CIL methods. On the other sides, popular rehearsal-based methods in the CIL problem cannot be applied in the FSCIL problem because it might result in all samples to be stored in the memory.  

In the earlier stages,  various FSCIL methods have been developed, offering different approaches i.e. backbone tuning \cite{F2M_shi2021overcoming}, meta-learning \cite{METAFSCIL_chi2022metafscil}, prototype-tuning \cite{IDLVQ_chen2021incremental}, and dynamic network approach \cite{CEC_zhang2021few}. Most of the methods fully train the backbone parameters in the base task, then train a small subset of the backbone or part of the model e.g. projection and weighting in the few-shot tasks. This implies that the model relies too much on the base task performance. In addition, the model barely achieves plasticity in the new tasks, while suffering from a high risk of forgetting on the base task.

Meanwhile, the use of the foundation model and the parameter-efficient fine-tuning (PEFT) approach offer a breakthrough solution to the catastrophic forgetting problem \cite{L2P_wang2022learning}. The pre-trained model that has strong background knowledge handles the dependency on the base task. In addition, the approach offers far smaller learnable parameters than the previous approaches and reduces the model training time by a large margin. The second advancement is the language-guided learning approach pioneered by \cite{LGCL_khan2023introducing} that contributes to the advancement of the FSCIL method. Language-guided learning offers discriminative representations through encoded text embedding of respective classes by a pre-trained vision-language model (VLM) such as CLIP\cite{CLIP_radford2021learning}. That is, the use of language modality alleviates the data scarcity problem because it offers complementary information of those input images.  

\begin{figure*}[h!]
\centering
\setlength{\abovecaptionskip}{-3pt plus 0pt minus 0pt}
\setlength{\belowcaptionskip}{-17pt plus 0pt minus 0pt}
\begin{center}
\includegraphics[width=1.0\textwidth]{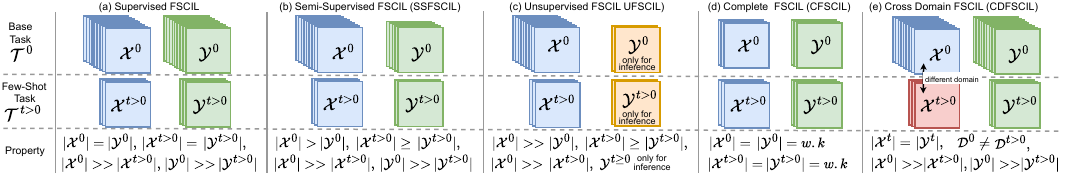}
\end{center}
\caption{\small \textcolor{black}{Types of FSCIL settings and their properties (a) Fully-Supervised, (b) Semi-Supervised (c) Unsupervised (d) Complete (e) Cross-Domain FSCIL}}
\label{fig:fscil_settings}
\end{figure*}

The recent survey \cite{NNSurvey_tian2024survey} systematically discussed the four approaches of FSCIL methods from the theoretical perspective and application. However, the survey doesn't cover the latest advancement of the PEFT approach and language-guided learning of FSCIL. Second, the survey also doesn't yet cover the important advancements of FSCIL settings i.e. semi-supervised (SSFCIL)\cite{SSFSCIL_cui2021semi}, unsupervised (UFSCIL)\cite{UNISA_10639442}, complete (CFSCIL)\cite{M2SD_lin2024m2sd} and cross-domain (CDFSCIL)\cite{MEDCDFSCIL_yang2023few} that have been applied in various research areas. Third, prototype bias and rectification\cite{PROTORECT_liu2020prototype} that are the important aspects of data scarcity haven't been discussed yet in the survey. Last but not least, the subtle details such as backbone and parameters, details of base classes, and novel classes performance i.e., stability-plasticity performance haven't been discussed as well. These intriguing factors motivate us to write a comprehensive survey on FSCIL. We offer a comprehensive FSCIL (sub-)settings, detailed topology, and formal objectives of each FSCIL approach, broad yet detailed aspects in FSCIL methods comparison, accommodating the latest advancement of FSCIL. Then, we present the open problem, potential solution and future direction of FSCIL. Specifically, our contributions are summed up: 
\vspace{-3pt}
\begin{itemize}[leftmargin=*]
\item We propose a comprehensive FSCIL topology highlighting the latest advancement of FSCIL methods, especially the new PEFT approach, the formal objective of each approach, and sub-settings of FSCIL i.e. SSFSCIL, UFSCIL, CFSCIL, and CDFSCIL.  
\vspace{-3pt}
\item We analyze broader aspects of FSCIL methods i.e. backbone type, learnable parameters, classifiers, the use of the pre-trained model (PTM), augmentation, general performance and stability-plasticity performance detailed in base classes, and novel classes performance.  
\vspace{-3pt}
\item We highlight the importance of prototype rectification to handle prototype bias due to data scarcity issues in the few-shot tasks. We classify the rectification mechanisms into a topology for a better understanding.
\vspace{-3pt}
\item We present a rigorous analysis of the current open problem, potential solution, and future direction of FSCIL related to applicable real-world constraints i.e. data privacy, data streams, and open-world problem.
\vspace{-3pt}
\end{itemize}

\section{Preliminaries}\label{preliminaries}
\subsection{Problem Formulation}\label{problem_formulation}
\textbf{Few-Shot Class-Incremental Learning (FSCIL)} is defined as:  Given 
a sequence of tasks $[0, 1, 2, ..., T]$ where each task $t$ carries a labeled training set $\mathcal{T}^{t}=\{(x_{i}^{t},y_{i}^{t})\}_{i=1}^{|\mathcal{T}^{t}|}$, where $x_i^t \in \mathcal{X}^t$ denotes an input image and $y_i^t \in \mathcal{Y}^t$ denotes its label, and $|.|$ denotes the cardinality. Each task $t$ is disjoint with another task $t'$. i.e. $\forall_{t,t'} \mathcal{T}^{t} \cap \mathcal{T}^{t'} = \emptyset$. The first task $(t=0)$ carries abundant training samples while the remaining tasks $(t>0)$ carries only far smaller samples than the first task i.e  $|\mathcal{T}^{0}| >> |\mathcal{T}^{t>0}|$. For a convenient way, task-$0$ is called the base task while the rest is called the few-shot task (FS task).
Let a deep neural network $g_{\phi}(f_{\theta}(.))$ be parameterized by $\theta$ and $\phi$ where $f(.)$ and $g(.)$ are the feature extractor and classifier respectively. \textbf{The objective of FSCIL} is to achieve the optimum parameters ${\theta*}$ and ${\phi*}$ so that $g_{\phi*}(f_{\theta*}(.))$ recognizes all the learned classes from the first task until the current task $t$ i.e. $\{\mathcal{T}^0,...,\mathcal{T}^t\}$. Formally, FSCIL objective is defined as in equation \ref{eq:fscil_obj}.
\vspace{-3pt}
\begin{equation}\label{eq:fscil_obj}
   \small
   \theta*, \phi*  = \argmin_{\theta, \phi} \mathbf{E}_{(x,y) \in \mathcal{T}^0 \cup \mathcal{T}^1... \mathcal{T}^t} \mathbf{I}(y=g_{\phi}(f_{\theta}(x)))
\end{equation}
\noindent where $\mathbf{E}$ denotes the expectation, $\mathbf{I}$ denotes the binary function that outputs 1 if the condition is met and 0 otherwise. 

\subsection{Types of FSCIL Setings}\label{fscil_settings}
There are 3 types of FSCIL setting based on the training mechanism with respect to the number of images and labels, i.e. $|\mathcal{X}|$ and $|\mathcal{Y}|$ as follows:

\noindent \textbf{a). Supervised FSCIL} or just FSCIL \cite{TOPIC_tao2020few} utilizes the fully supervised training mechanism that is conducted in the case where the number of images and labels are the same for every task, i.e. $|\mathcal{X}^t| = |\mathcal{Y}^t|$ for all $t \in [1,2..T]$. In the few-shot task ($t > 0$), the number of images and labels are the same and equal to the $k$-shot value.

\noindent \textbf{b). Semi-Supervised FSCIL (SSFSCIL)}\cite{SSFSCIL_cui2021semi} deploys supervised and unsupervised training mechanisms in the case where the number of images is greater than the number of labels, i.e. $|\mathcal{X}^t| > |\mathcal{Y}^t|$ for any $t \in [1,2..T]$. The semi-supervised training is not necessarily conducted in all tasks, it can be conducted only in the base task where abundant samples are available but only fewer labels are provided. 

\noindent \textbf{c). Unsupervised FSCIL (UFSCIL)}\cite{UNISA_10639442} deploys only unsupervised training mechanisms to deal with the unavailability of training label i.e. $|\mathcal{Y}^t| = 0$ for all $t \in [1,2..T]$. A small fraction of labels are available i.e. $|\mathcal{Y}^t| << |\mathcal{X}^t|$ only for inference purpose since the model needs to give predicted label for any input $x$. 

\noindent \textbf{d). Complete FSCIL (CFSCIL)}\cite{CFSCIL_hersche2022constrained} remove the assumption that the base task has abundant labeled samples. Thus, all task only hold a few labeled samples and organized in n-way k-shot setting i.e. $|\mathcal{X}^t| = |\mathcal{Y}^t| = w.k$ for all $t \in [1,2..T]$, where $w$ is the number of classes in each task, and $k$ is the number of samples (shot) for each class. 

\noindent \textbf{e). Cross Domain FSCIL (CDFSCIL)}\cite{CFSCIL_hersche2022constrained} is similar to supervised FSCIL, but assumes that few-shot task and base task have different domains i.e. $\mathcal{D}^0 \neq \mathcal{D}^{t>0}$. The illustration and property of all settings is visualized in fig.\ref{fig:fscil_settings}. 

\section{FSCIL Approaches}
\begin{figure*}
\setlength{\abovecaptionskip}{-5pt plus 0pt minus 0pt}
\setlength{\belowcaptionskip}{-15pt plus 0pt minus 0pt}
\begin{center}
\includegraphics[width=0.999\textwidth]{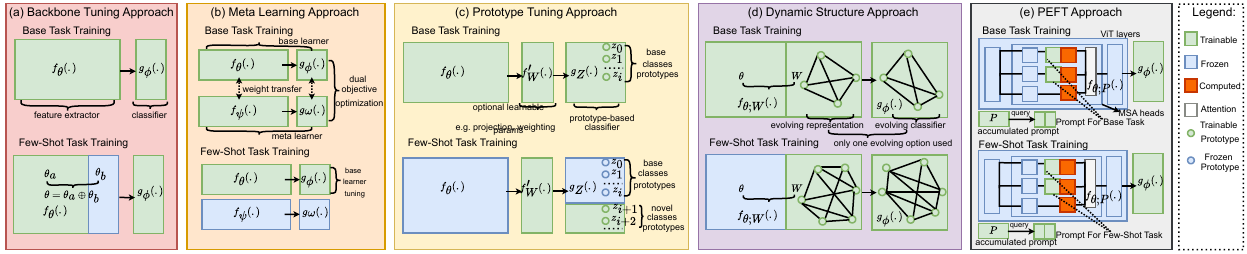}
\end{center}
\caption{\small \textcolor{black}{Visualization of FSCIL Approaches i.e. (a) backbone tuning approach, (b) meta learning approach, (c) prototype tuning approach, (d) dynamic structure approach, and (e) parameter efficient fine tuning (PEFT) approach.}}
\label{fig:typeof_approaches}
\end{figure*}

\subsection{Backbone Tuning Approach}\label{backbone_approach}
The backbone tuning approach trains the whole backbone and classifier parameters in the base task, then trains the classifier and some of backbone parameters in the few-shot tasks as illustrated in fig. \ref{fig:typeof_approaches} (a). Suppose that $\theta = \theta_a \oplus \theta_b$, where $\theta_a$ and $\theta_b$ are front part (earlier layers) and back part (latest layers) of the feature extractor parameters, and $\oplus$ denotes concatenation. In the base task, the method optimizes $\theta$ and $\phi$, while in the few-shot tasks, the method freezes $\theta_a$ and optimizes $\theta_b$ and $\phi$. In general, the objective of backbone tuning approach is defined as in equation \ref{eq:backbone_tuning_obj}.
\vspace{-3pt}
\begin{equation}\label{eq:backbone_tuning_obj}
   \small
   \begin{split}
    \theta*, \phi*  = \argmin_{\theta, \phi} \mathbf{E}_{(x,y) \in \mathcal{T}^0} \mathbf{I}(y=g_{\phi}(f_{\theta}(x))) \text{ if } (t=0) \\
    \theta_b*, \phi*  = \argmin_{\theta_b, \phi} \mathbf{E}_{(x,y) \in \mathcal{T}^0 \cup ... \mathcal{T}^t} \mathbf{I}(y=g_{\phi}(f_{\theta}(x))) \text{ else } 
   \end{split}
\end{equation}

Several strategies are enforced in this aproaches i.e. GPTree \cite{GPTREE_achituve2021gp} utilizes Gaussian Process to approximate the distribution of classes with few-shot labeled samples. Similarly, LDC\cite{LDC_liu2023learnable} enforces a learnable models to approximate the distribution. F2M\cite{F2M_shi2021overcoming} and FLOWER\cite{FLOWER_10571377} create a flat loss region by perturbating latest part of the backbone parameters i.e. $\mathcal{L}((x,y);\theta_a\oplus (\theta_b+\epsilon);\phi))$ is minimal for any $-b \leq \epsilon \leq b$, where $\epsilon$ and $b$ is perturbation noise and bound respectively. ERDR\cite{ERDR_liu2022few} and \cite{SUPSPACE} utilize customized regularization strategy i.e. entropy and subspace regularization respectively. FeSSS\cite{FESSS_ahmad2022few} utilizes self-supervised method and feature fusion to obtain more discrimivative features. Mixed-up features are also utilized by M2SD\cite{M2SD_lin2024m2sd} and combined with mixed-up distillations.  SoftNet\cite{SOFTNET_kang2022soft} adds binary mask $m$ and gating function to sub-network of $\theta$. It  enables different weight of $\theta$ for different samples. On the other hand, WaRP\cite{WARP_kim2023warping} transforms sub-net of $\theta$ with multiple axis rotation to gain distinctive representations. 

\subsection{Meta-Learning Approach} \label{meta_approach}
\vspace{-2pt}
Inspired from MAML\cite{MAML_finn2017model} meta learning approach (fig. \ref{fig:typeof_approaches} (b)) trains both meta learner and base learner with the different sets of data i.e. training and validation set. Validation set can be obtained from auxiliary (open) dataset or by transforming the training set with sequence of operations. Suppose that $f_{\psi}(.)$ is the meta learner parametrized by $\psi$, and $\mathcal{T}_{v}^t = \text{ Transform}(\mathcal{T}^t)$ is validation set transformed from $\mathcal{T}^t$. then the dual objective of meta learning-based FSCIL is formulated in eq. \ref{eq:meta_learning_obj}.
\vspace{-3pt}
\begin{equation}\label{eq:meta_learning_obj}
   \small
   \begin{split}
    \psi*  = \argmin_{\psi} \mathbf{E}_{(x,y) \in \mathcal{T}_v^0 \cup ... \mathcal{T}_v^t} \mathcal{L}(f_{\psi}(x))  \\
   \text{s.t. } \theta*, \phi*  = \argmin_{\theta, \phi} \mathbf{E}_{(x,y) \in \mathcal{T}^0 \cup ... \mathcal{T}^t} \mathbf{I}(y=g_{\phi}(f_{\theta}(x)))
   \end{split}
\end{equation}
MetaFSCIL\cite{METAFSCIL_chi2022metafscil} utilizes bidirectional guided modulation (BGM) that allows meta and base learners to map their weights one to another. C-FSCIL\cite{CFSCIL_hersche2022constrained} utilize meta learning to perform 3 modes of prototypes training i.e. averaged prototypes,  retraining on bipolarized-prototypes, and retraining on nudged-prototypes. Both MetaFSCIL and C-FSCIL train meta learner only in the base task, it is frozen afterwards. XTarNet\cite{XTARNET_yoon2020xtarnet} deploys CNN as meta learner, utilizes a pretrained backbone, and adds two additional modules i.e. MergeNet and TConNet that performs weights merger and conditioning classifier respectively. Similar mechanism is performed by LIMIT\cite{LIMIT_zhou2022few} that utilize meta learning for prototype tuning. LIMIT generates multiple prototypes from an input instance, then the prototypes are calibrated by meta learner then aggregated with inner product into a final prototype.

\subsection{Prototype-Tuning Approach}\label{proto_approach}
Prototype tuning approach is the most common method in FSCIL due to its simplicity and scalability. Suppose that $z_c$ is the prototype of class $c$, $Z^t={z^t_c}$ is the prototype set of task $t$ contains prototypes of any class $c$ exists in $\mathcal{T}^t$ , and $Z = Z^0 \cup Z^1, ... \cup Z^t$ is the prototype set of all learned classes from task $0$ to task $t$. The approach utilizes prototype-based classifier $g_Z(.)$ instead of standard MLP classifier  $g_{\phi}(.)$. Thus it optimizes $Z$ in every task, while feature extractor parameter $\theta$ is optimized only in the base task. A prototype-based method may utilize an additional network $f'_W(.)$ parameterized by $W$ between backbone $f_{\theta}(.)$ and prototype-based classifier $g_Z(.)$ for projection, weighting, or refinement. The objective of the prototype tuning approach is defined in eq. \ref{eq:backbone_tuning_obj}.
\vspace{-3pt}
\begin{equation}\label{eq:proto_tuning_obj} 
   \scriptsize
   \begin{split}
    \theta*, W*, Z*  = \argmin_{\theta, W, Z} \mathbf{E}_{(x,y) \in \mathcal{T}^0} \mathbf{I}(y=g_{Z}(f'_W(f_{\theta}(x))))  \text{ if} (t=0) \\
    W*, Z*  = \argmin_{Z} \mathbf{E}_{(x,y) \in \mathcal{T}^0 \cup ... \mathcal{T}^t} \mathbf{I}(y=g_{Z}(f'_W(f_{\theta}(x))))  \text{ else}
   \end{split}
\end{equation}
IDLVQ\cite{IDLVQ_chen2021incremental} optimizes vector $Z$ to directly classify backbone feature $f_{\theta}(x)$. There are several variations to implement $f'_W(.)$ i.e. dynamic relation projection as in SPRR\cite{SPRR_zhu2021self}, subspace projection as in SFMS\cite{SFMS_cheraghian2021synthesized}, orthogonal projection as in ORCO\cite{ORCO_ahmed2024orco},  merged multiple embeddings by attention as in SAKD\cite{SAKD_cheraghian2021semantic}, kernel analytics module as in GKEAL\cite{GKEAL_zhuang2023gkeal}, positive-negative weighting matrices as in CLOM\cite{CLOM_zou2022margin}, compositional weighting as in Comp-FSCIL\cite{COMP_FSCIL_zou2024compositional}, prototype rectification as in TEEN \cite{TEEN_wang2024few} and YOURSELF\cite{YOURSELF_tang2024rethinking} or any customized refinement.  Along with prototype tuning, this approach also applies conceptual learning methods e.g. contrastive learning as in SAVC\cite{SAVC_song2023learning}, virtual(augmented) prototypes as in FACT\cite{FACT_zhou2022forward} and ALICE\cite{ALICE_peng2022few}, neural collapse as in NC-FSCIL\cite{NC_FSCIL_yang2023neural}, and Slow-and-Fast learning as in MgSvf\cite{MGSVF_zhao2021mgsvf}.  

\subsection{Dynamic Architecture Approach}\label{dynamic_arc}
Dynamic architecture approach tackles FSCIL challenges by incrementally evolving its architecture to adapt to the new tasks. Fig. \ref{fig:typeof_approaches} (d) shows that this approach can be implemented into two types of growing component,  i.e. growing representation $W$ or growing classifier $\phi$.  TOPIC\cite{TOPIC_tao2020few} grows neural gas $W$, a graph-like structure to obtain a more discriminative class representation. BiDist\cite{BIDIST_zhao2023few} adds learnable weight for each task $W^t$ then uses bilateral distillation of current and previous tasks representation.    CEC\cite{CEC_zhang2021few} transforms incrementally added linear classifier into graph structure. S3C grows stochastic classifiers by continually adding 4-angles (0,90,180,270)  representations.  LEC-Net\cite{LECNET_yang2021learnable} expands learnable compression network $W$ to mitigate overfitting due to data scarcity. Similarly, DSN\cite{DSN_yang2022dynamic} expands self-activation and compression network.
The objective of the dynamic architecture approach is defined in eq. \ref{eq:dynamic_arch}.
\vspace{-5pt}
\begin{equation}\label{eq:dynamic_arch}
   \scriptsize
   \begin{split}
    \theta*, W*, \phi*  = \argmin_{\theta, W, \phi} \mathbf{E}_{(x,y) \in \mathcal{T}^0} \mathbf{I}(y=g_{\phi}(f_{\theta;W}(x))) \text{ if } (t=0) \\
    W*, \phi*  = \argmin_{W, \phi} \mathbf{E}_{(x,y) \in \mathcal{T}^0 \cup ... \mathcal{T}^t} \mathbf{I}(y=g_{\phi}(f_{\theta;W}(x))) \text{ else } 
   \end{split}
\end{equation}

\vspace{-5pt}
\subsection{Parameter Efficient Fine Tuning (PEFT) Approach}\label{peft_approach}
The PEFT approach proposes a unique new solution for FSCIL by training small-sized parameters (\textbf{prompts}) on top of frozen pre-trained backbone e.g. visual transformer (ViT). 
Different to CNN, ViT has sequence of layers, where a layer contains Multiples Self Attention (MSA) heads combined with other components, each performing self-attention function as in eq. \ref{eqatt}:
\vspace{-5pt}
\begin{equation}\label{eqatt}
    \small 
   A(Q, K, V) = Softmax((Q.K^T)/\sqrt{D/h})V
\end{equation}
where $Q, K$, and $V$ are query, key, and value vectors obtained by processing an input patch into respective weights i.e.  $x.W^Q, x.W^K$, and $x.W^V$, $D$ and $h$ denote patch dimension and number of heads respectively. Prompt-based method attaches a small learnable prompts into a frozen pre-trained ViT model (fixed $\psi$) \cite{L2P_wang2022learning}. The attachment techniques are mainly divided into 2 types i.e. prompt tuning and prefix tuning \cite{PREFIXTUN_li2021prefix}. Prompt tuning as in L2P \cite{L2P_wang2022learning} inserts a prompt vector $P$ into query, key, and value of MSA, then the attention mechanism is computed as $A([P\oplus Q], [P \oplus K], [P \oplus V])$, where $\oplus$ denotes concatenation operation. Prefix tuning as in DualP \cite{DualP_wang2022dualprompt} divides $P$ into pair of $P^K$ and $P^V$ for key and value respectively, then the attention mechanism becomes $A(Q, [P^K \oplus K], [P^V \oplus V])$. The structure of $P$ can be defined by several options, i.e. pool-based structure as in L2P, task-wise structure as in DualP, growing component structure as in CODA-P\cite{CODA_smith2023coda}, or any customized structure. 

ASP\cite{ASP_liu2024few} deploys task-aware task-invariant prompts to improve prompt selection during the testing phase. Besides, it utilizes a prompt encoder that produces an additional (second) prompt. Privilege \cite{PRIVILEGE_park2024pre} leverages vision and language modalities and prepends concatenation of vision and language prompts. Similarly, FineFMPL\cite{FINEFMPL_sunfinefmpl} performs language-guided prompting i.e. tuning vision prototype by language prototypes. In addition, it utilizes two types of vision prototypes i.e. global level and object level prototypes. ApproxFSCIL\cite{APPROXFSCIL_wang2025approximation} concatenates ViT layers with norm layers and tanh function instead of prompt, then bounds training process with transfer risk and consistency risk. Formally, the objectives of PEFT approach is presented in eq.\ref{eq:peft_obj}.
\vspace{-2pt}
\begin{equation}\label{eq:peft_obj} 
   \small
    P*, \phi*= \argmin_{P, \phi} \mathbf{E}_{(x,y) \in  \mathcal{T}^0 \cup ... \mathcal{T}^t} \mathbf{I}(y=g_{\phi}(f_{\theta;P}(x)))  
\end{equation}

\begin{figure*}
\setlength{\abovecaptionskip}{-7pt plus 0pt minus 0pt}
\setlength{\belowcaptionskip}{-7pt plus 0pt minus 0pt}
\begin{center}
\includegraphics[width=0.99\textwidth]{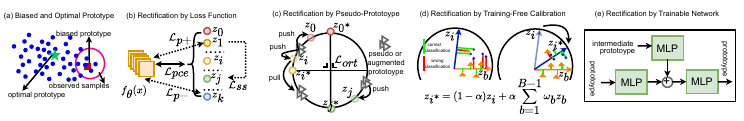}
\end{center}
\caption{\small \textcolor{black}{Biased Prototype, Optimal Prototypes, and Types of Prototype Rectification Mechanism (a) Biased and Optimal Prototype (b) Rectification by Loss function only (c) by pseudo-prototypes (d) by training-free calibration (e) by trainable networks}}
\label{fig:proto_rect}
\end{figure*}

\vspace{-9pt}
\section{Trends and Finding}
\subsection{Backbone, Parameters, and Pre-trained Model (PTM)}
As shown in tab. \ref{tab:main_comparison}, ResNet18 is the preferred backbone for backbone tuning, meta-learning, prototype-tuning, and dynamic structure approach following the pioneer of FSCIL setting\cite{TOPIC_tao2020few}. However, the recent advancement of PEFT approach utilized ViT instead of ResNet. This is one of the clear impacts of the advancement of the foundation model. Despite utilizing a bigger backbone i.e. ViT-B/16 with 86M parameters, PEFT approach has the smaller trainable parameter i.e. 0.2-11.9M as all the pre-trained backbone weights are frozen, except YourSelf. YourSelf fine-tunes its backbone in the base classes resulting a bigger trainable parameters, albeit utilizing fewer layers of ViT. 

The other 4 approaches have bigger trainable parameters i.e. 11.7-41.7M albeit utilizing the smaller backbone i.e. 11.7M as these approaches tune the whole backbone in the base task. Note that a method usually has more parameters than the backbone and classifiers e.g. projection, weighting, self-attention, or rectifier. Thus, these approaches train more parameters. PEFT utilizes a PTM for all scenarios, while the rest utilizes PTM in the CUB dataset. This fact gives us 2 insights, first, training the backbone from scratch isn't enough for a dataset with small (30 or fewer) samples per class, and second we can compare all the methods head to head in CUB dataset as all the methods utilize PTM.

\subsection{Prototype-based over Network-based Classifiers} 
The table shows that the prototype-based classifier is highly preferable than the network-based classifier such as MLP or reconstructed linear network \cite{CEC_zhang2021few}. A prototype-based classifier uses classes' prototypes where a prototype is optimized only in a particular task. Thus, it won't be adjusted in the upcoming tasks. Thus, it reduces the risk of forgetting. On the contrary, a network-based classifier is adjusted in each task, thus it has a higher chance of forgetting. 

Several methods utilize combined or customized classifiers. GKEAL and Comp-FSCIL utilize a combination of linear and prototype-based classifiers. The methods have a selection mechanism to choose which classifiers will be utilized in the inference phase. PriViLege and FineMPL utilize both vision and language prototypes since the methods adopt language-guided learning. S3C generates multiple prototypes with angular rotation, and then performs a stochastic classification based on the similarity to the prototypes.   

\subsection{Image vs Feature Augmentation} 
Several FSCIL methods utilize augmentation to handle data scarcity in FSCIL, aside from standard augmentation such as random crop and random flip, commonly used in a data loader setting. Image augmentation offers a low-level and simple procedure but consumes higher system memory. On the contrary, augmentation in the feature space requires a lower memory size, but it needs more complex operation and validation.  For example, image-space augmentation can be validated via a visual plot, while feature-space augmentation should be evaluated by valid metrics such as Euclidean distance, or loss. GP-tree augments its likelihood estimator rather than image or prototype augmentations since it performs a Gaussian process in its training phase. Tab.\ref{tab:main_comparison} shows that the latest methods don't perform any augmentations.

\vspace{-5pt}
\subsection{The Importance of Prototype Rectification}
As shown in tab.\ref{tab:main_comparison} all the methods that utilize prototype-based classifiers perform prototype rectification to maximize class discrepancies. As emphasized in \cite{PROTORECT_liu2020prototype}, and visualized in fig.\ref{fig:proto_rect}(a) a few observed samples have a high chance of producing a biased prototype i.e. a prototype that only represents the few observed samples but not the whole distribution. Thus a rectification is necessary to adjust the biased prototype into the optimal prototype. 

The existing FSCIL methods perform various types of rectification as illustrated in fig.\ref{fig:proto_rect}(b)-(d). The simplest mechanism is rectification by loss where classes' prototypes are adjusted based on similarity to the input features or between classes prototypes.  There are many types of loss that can be utilized to perform this rectification e.g. prototypical-cross-entropy $\mathcal{L}_{pce}$, self-supervised loss $\mathcal{L}_{ss}$, or contrastive loss based on positive and negative prototypes i.e. $\mathcal{L}_{p+}$ and $\mathcal{L}_{p-}$. The second type is rectification by pseudo prototypes. Pseudo-prototypes can be obtained by generating several prototypes from different layers, model perturbations, or augmentation. Some methods like OrCo\cite{ORCO_ahmed2024orco} utilize orthogonal loss $\mathcal{L}_{ort}$ to minimize dot product between classes' prototypes. Rectification by training-free calibration \cite{TEEN_wang2024few} adjusts a prototype of novel classes based on weighted similarities to the base classes' prototypes. While rectification by trainable networks utilizes an extra trainable model to calibrate the prototype \cite{YOURSELF_tang2024rethinking}. Training-free calibration is less complex than rectification by pseudo-prototypes or trainable networks since it doesn't need extra operations for generating pseudo-prototypes.

\subsection{The Use of Language-Guided Learning}
\vspace{-2pt}
The advancement of the vision-language model (VLM) such as CLIP\cite{CLIP_radford2021learning} drives the utilization of language-guided learning in FSCIL. PriViLege\cite{PRIVILEGE_park2024pre} deploys language prototypes along with vision prototypes as loss anchors and classifiers, while FineMPL\cite{FINEFMPL_sunfinefmpl} merges both prototypes into one long prototype and then uses it for the same purpose. Despite only 2 methods that utilize language-guided learning, the studies show a promising future for the approach.

\subsection{Performance} 
\vspace{-2pt}
Tab.\ref{tab:main_comparison} shows the overall performance on 3 main FSCIL benchmark datasets, in terms of average accuracy(AA) from the first until the last task, performance drop (PD) between the first and the last task and average harmonic accuracy (AHM) that represent stability-plasticity balance from the second until the last task. In general, AA in three datasets that is still below 90\% shows that FSCIL offers a big opportunity for improvement. The high accuracy ($>$ 90\%) in MiniImageNet by Privilege, FineMPL and ApproxFSCIL are common-sense results as the methods utilize pre-trained ViT on ImageNet (the superset of MiniImageNet). The results on the CUB dataset, where the PEFT method achieves relatively higher AA show the impact of the bigger backbone i.e. ViT (86.7M) over ResNet18(11.7M) since all methods utilize PTM. The high magnitude of PD i.e. up to 42\% shows that catastrophic forgetting with data scarcity remains a difficult problem. 

PEFT methods e.g. ASP and Privilege confirm the superiority of the PEFT approach to the other 4 approaches with a significant gap i.e. up to 20\%. The improvement from L2P, DualP, and CodaP to L2P+, DualP+, and CodaP+ confirms the significant impact of the prototype-based classifier for FSCIL methods.  The other 4 approaches have competitive performance in those 3 datasets, with similar trends i.e. the latest methods outperform the earlier methods. Both from AA and PD measurement,  the best performers of those 4 approaches are below the best performer of the PEFT approach.

\begin{table*}[h!]
\centering
\tiny
\setlength{\tabcolsep}{0.45em}
\begin{tabular}{lccccccccccccccccccc}
\hline
 &  &  & \multicolumn{2}{c}{\textbf{BackBone}} &  &  &  &  &  &  & \multicolumn{3}{c}{\textbf{Perf. On MiniImageNet}} & \multicolumn{3}{c}{\textbf{Perf. On CIFAR100}} & \multicolumn{3}{c}{\textbf{Perf. On CUB}} \\ \cline{4-5} \cline{12-20}
\multirow{-2}{*}{\textbf{Method}} & \multirow{-2}{*}{\textbf{Year}} & \multirow{-2}{*}{\textbf{Venue}} & \textbf{Type} & \textbf{Size} & \multirow{-2}{*}{\textbf{\begin{tabular}[c]{@{}c@{}}Learn.\\ Param.\end{tabular}}} & \multirow{-2}{*}{\textbf{\begin{tabular}[c]{@{}c@{}}Utilize\\ PTM on\end{tabular}}} & \multirow{-2}{*}{\textbf{Classifier}} & \multirow{-2}{*}{\textbf{Aug}} & \multirow{-2}{*}{\textbf{\begin{tabular}[c]{@{}c@{}}Proto.\\ Rect.\end{tabular}}} & \multirow{-2}{*}{\textbf{\begin{tabular}[c]{@{}c@{}}Lang.\\ Guided\end{tabular}}} & \textbf{AA} & \textbf{PD} & \textbf{AHM} & \textbf{AA} & \textbf{PD} & \textbf{AHM} & \textbf{AA} & \textbf{PD} & \textbf{AHM} \\ \hline
\cellcolor[HTML]{EC9BA4}GP-Tree & 2021 & ICML & ResNet18 & 11.67M & - & CUB & Network & Likelihood & \xmark & \xmark & 46.68 & 14.10 & - & - & - & - & 54.26 & 30.12 & - \\
\cellcolor[HTML]{EC9BA4}FSLL & 2021 & AAAI & ResNet18 & 11.67M & - & CUB & Prototype & \xmark & Loss & \xmark & 52.58 & 13.90 & - & 47.59 & 16.51 & - & 60.93 & 18.56 & - \\
\cellcolor[HTML]{EC9BA4}F2M & 2021 & NIPS & ResNet18 & 11.67M & 11.67M & CUB & Prototype & \xmark & Loss & \xmark & 54.89 & 12.39 & - & 53.65 & 11.06 & - & 69.49 & 20.81 & - \\
\cellcolor[HTML]{EC9BA4}ERDR & 2022 & ECCV & ResNet18 & 11.67M & - & CUB & Network & \xmark & \xmark & \xmark & 58.02 & 13.82 & - & 60.78 & 13.62 & - & 61.52 & 23.51 & - \\
\cellcolor[HTML]{EC9BA4}SubspaceReg. & 2022 & ICLR & ResNet18 & 11.67M & 26.30M* & - & Network & \xmark & \xmark & \xmark & 63.71 & 16.66 & - & - & - & - & - & - & - \\
\cellcolor[HTML]{EC9BA4}FeSSSS & 2022 & CVPR & ResNet18 & 11.67M & - & CUB & Prototype & \xmark & Loss & \xmark & 68.24 & 13.26 & - & - & - & - & 62.86 & 26.62 & - \\
\cellcolor[HTML]{EC9BA4}LDC & 2023 & TPAMI & ResNet18 & 11.67M & - & CUB & Prototype & Feature & Tr. Net. & \xmark & 63.99 & 11.83 & - & 64.35 & 11.61 & - & 68.32 & 16.31 & - \\
\cellcolor[HTML]{EC9BA4}SoftNet & 2023 & ICLR & ResNet18 & 11.67M & 22.38M* & CUB & Prototype & \xmark & Loss & \xmark & 64.32 & 13.78 & - & 65.86 & 14.14 & - & 64.74 & 22.00 & - \\
\cellcolor[HTML]{EC9BA4}WaRP & 2023 & ICLR & ResNet18 & 11.67M & - & CUB & Network & \xmark & \xmark & \xmark & 74.12 & 9.18 & - & 74.55 & 11.65 & - & 55.86 & 18.38 & - \\
\cellcolor[HTML]{EC9BA4}FLOWER & 2024 & TNNLS & ResNet18 & 11.67M & 12.25M & CUB & Prototype & Feature & Loss & \xmark & 59.55 & 15.05 & - & 59.42 & 13.98 & - & 65.67 & 21.53 & - \\
\cellcolor[HTML]{EC9BA4}M2SD & 2024 & AAAI & ResNet18 & 11.67M & - & CUB & Network & Image & \xmark & \xmark & 68.63 & 13.48 & - & 68.25 & 13.55 & - & 68.24 & 19.66 & - \\ \hline
\cellcolor[HTML]{FFB66C}XtarNet & 2021 & ICML & Resnet18 & 11.67M & - & - & Prototype & \xmark & Loss & \xmark & 66.86 & - & - & - & - & - & - & - & - \\
\cellcolor[HTML]{FFB66C}C-FSCIL & 2022 & CVPR & ResNet18 & 11.67M & 12.79M* & - & Prototype & \xmark & Loss & \xmark & 61.61 & 14.79 & 29.55 & 61.64 & 15.83 & 25.66 & - & - & - \\
\cellcolor[HTML]{FFB66C}Meta-FSCIL & 2022 & CVPR & ResNet18 & 11.67M & - & CUB & Network & \xmark & \xmark & \xmark & 58.85 & 13.19 & - & 60.79 & 13.71 & - & 61.93 & 23.26 & - \\
\cellcolor[HTML]{FFB66C}LIMIT & 2022 & TPAMI & ResNet18 & 11.67M & 12.33M & CUB & Prototype & Feature & Loss & \xmark & 59.06 & 13.26 & 33.36 & 61.84 & 11.97 & 33.89 & 66.67 & 17.87 & 53.57 \\ \hline
\cellcolor[HTML]{FFE994}SPPR & 2021 & CVPR & ResNet18 & 11.67M & 12.36M & CUB & Prototype & \xmark & Loss & \xmark & 52.76 & 8.69 & - & 54.52 & 9.58 & - & 49.32 & 31.35 & - \\
\cellcolor[HTML]{FFE994}IDLVQ & 2021 & ICLR & ResNet18 & 11.67M & - & CUB & Prototype & \xmark & Loss & \xmark & 51.16 & 13.61 & - & 54.89 & 12.39 & - & 65.23 & 19.56 & - \\
\cellcolor[HTML]{FFE994}SAKD & 2021 & CVPR & ResNet18 & 11.67M & - & CUB & Prototype & \xmark & Pse. Proto. & \xmark & 50.86 & 11.14 & - & 50.63 & 10.77 & - & 46.13 & 35.27 & - \\
\cellcolor[HTML]{FFE994}SFMS & 2021 & ICCV & ResNet18 & 11.67M & - & CUB & Prototype & Feature & Pse. Proto. & \xmark & 50.77 & 10.60 & - & 50.86 & 11.17 & - & 51.84 & 25.55 & - \\
\cellcolor[HTML]{FFE994}FACT & 2022 & CVPR & ResNet18 & 11.67M & 11.34M & CUB & Prototype & Feature & Pse. Proto. & \xmark & 60.79 & 15.51 & 24.16 & 63.10 & 15.40 & 34.91 & 64.42 & 18.96 & 51.58 \\
\cellcolor[HTML]{FFE994}ALICE & 2022 & ECCV & ResNet18 & 11.67M & 41.71M & CUB & Prototype & Image & Loss & \xmark & 63.99 & 16.61 & - & 63.21 & 15.79 & - & 65.75 & 17.30 & - \\
\cellcolor[HTML]{FFE994}MgSvF & 2022 & TPAMI & ResNet18 & 11.67M & - & CUB & Prototype & \xmark & Loss & \xmark & 52.65 & 10.24 & - & 61.32 & 12.74 & - & 62.37 & 17.96 & - \\
\cellcolor[HTML]{FFE994}CLOM & 2022 & CVPR & ResNet18 & 11.67M & 18.89M & CUB & Network & \xmark & Loss & \xmark & 57.22 & 14.88 & - & 60.34 & 13.86 & - & 65.48 & 22.60 & - \\
\cellcolor[HTML]{FFE994}NC-FSCIL & 2023 & ICLR & ResNet18 & 11.67M & 15.91M & CUB & Network & \xmark & Loss & \xmark & 67.82 & 16.20 & 52.62 & 67.50 & 15.02 & 47.90 & 67.28 & 21.01 & 55.03 \\
\cellcolor[HTML]{FFE994}GKEAL & 2023 & CVPR & ResNet18 & 11.67M & - & CUB & Compositional & Feature & Loss & \xmark & 60.48 & 13.11 & - & 61.35 & 12.66 & - & 66.35 & 20.21 & - \\
\cellcolor[HTML]{FFE994}TEEN & 2023 & NIPS & ResNet18 & 11.67M & - & CUB & Prototype & \xmark & TF Calib. & \xmark & 61.44 & 12.09 & 31.67 & 63.10 & 11.82 & 39.81 & 66.63 & 17.95 & 53.67\textasciicircum{} \\
\cellcolor[HTML]{FFE994}SAVC & 2023 & CVPR & ResNet18 & 11.67M & 24.29M & CUB & Prototype & Feature & Pse. Proto. & \xmark & 66.66 & 14.34 & - & 62.76 & 15.64 & - & 69.05 & 18.90 & - \\
\cellcolor[HTML]{FFE994}ORCO & 2024 & CVPR & ResNet18 & 11.67M & - & CUB & Prototype & Feature & Pse. Proto. & \xmark & 67.13 & 16.17 & 58.11 & 62.10 & 17.98 & 57.11 & 62.41 & 17.66 & 56.63 \\
\cellcolor[HTML]{FFE994}Comp-FSCIL & 2024 & ICML & ResNet18 & 11.67M & - & CUB & Compositional & \xmark & Loss & \xmark & 68.84 & 13.94 & - & 67.08 & 13.85 & - & 68.47 & 19.77 & - \\
\cellcolor[HTML]{FFE994}CLOSER & 2024 & ECCV & ResNet18 & 11.67M & - & CUB & Prototype & \xmark & Loss & \xmark & 62.78 & 13.24 & - & 63.10 & 12.62 & - & 69.07 & 15.82 & - \\ \hline
\cellcolor[HTML]{B7B3CA}TOPIC & 2020 & CVPR & ResNet18 & 11.67M & - & CUB & Network & \xmark & \xmark & \xmark & 39.64 & 21.67 & - & 42.62 & 21.48 & - & 43.92 & 42.40 & - \\
\cellcolor[HTML]{B7B3CA}CEC & 2021 & CVPR & ResNet18 & 11.67M & 12.34M & CUB & Network & \xmark & \xmark & \xmark & 57.75 & 14.25 & 29.35 & 59.53 & 13.54 & 35.13 & 61.33 & 23.57 & 46.01 \\
\cellcolor[HTML]{B7B3CA}LEC-net & 2022 & Arxiv & ResNet18 & 11.67M & - & CUB & Network & \xmark & \xmark & \xmark & 37.93 & 23.38 & - & 43.14 & 20.96 & - & 45.09 & 38.90 & - \\
\cellcolor[HTML]{B7B3CA}DSN & 2022 & TPAMI & ResNet18 & 11.67M & - & CUB & Network & Feature & \xmark & \xmark & - & - & - & 60.14 & 12.86 & - & 63.32 & 21.90 & - \\
\cellcolor[HTML]{B7B3CA}S3C & 2022 & ECCV & ResNet18 & 11.67M & 13.06M & CUB & Stoc. Prototype & Image & Loss & \xmark & 62.74 & 14.26 & - & 64.51 & 13.49 & - & 66.42 & 21.90 & - \\
\cellcolor[HTML]{B7B3CA}BiDist & 2023 & CVPR & ResNet18 & 11.67M & 11.50M* & CUB & Prototype & \xmark & Loss & \xmark & 61.30 & 13.30 & 40.48 & 66.17 & 13.33 & 37.65 & 67.35 & 18.20 & 50.87 \\ \hline
\cellcolor[HTML]{CCCCCC}L2P & 2022 & CVPR & ViT-B/16 & 86.57M & 0.21M & All & Network & \xmark & \xmark & \xmark & 72.97 & 21.62 & - & 70.74 & 20.78 & 0.00 & 61.37 & 44.46 & 3.28\textasciicircum{} \\
\cellcolor[HTML]{CCCCCC}DualP & 2022 & ECCV & ViT-B/16 & 86.57M & 0.33M & All & Network & \xmark & \xmark & \xmark & 73.31 & 21.74 & - & 70.37 & 21.74 & 0.10 & 62.09 & 44.12 & 5.81\textasciicircum{} \\
\cellcolor[HTML]{CCCCCC}CodaP & 2023 & CVPR & ViT-B/16 & 86.57M & 0.61M & All & Network & \xmark & \xmark & \xmark & - & - & - & 71.98 & 21.42 & 0.00 & 62.97 & 43.78 & 6.42\textasciicircum{} \\
\cellcolor[HTML]{CCCCCC}L2P+ & 2024 & ECCV & ViT-B/16 & 86.57M & 0.21M & All & Prototype & \xmark & Loss & \xmark & - & - & - & 77.73 & 6.97 & 68.00 & 75.42 & 8.63 & 73.40\textasciicircum{} \\
\cellcolor[HTML]{CCCCCC}DualP+ & 2024 & ECCV & ViT-B/16 & 86.57M & 0.33M & All & Prototype & \xmark & Loss & \xmark & - & - & - & 81.16 & 4.84 & 75.30 & 77.56 & 6.64 & 76.57\textasciicircum{} \\
\cellcolor[HTML]{CCCCCC}CodaP+ & 2024 & ECCV & ViT-B/16 & 86.57M & 0.61M & All & Prototype & \xmark & Loss & \xmark & - & - & - & 79.86 & 6.14 & 72.20 & 73.95 & 6.96 & 72.80\textasciicircum{} \\
\cellcolor[HTML]{CCCCCC}ASP & 2024 & ECCV & ViT-B/16 & 86.57M & - & All & Prototype & \xmark & Loss & \xmark & - & - & - & 89.02 & 3.18 & 85.30 & 83.20 & 3.98 & 83.56\textasciicircum{} \\
\cellcolor[HTML]{CCCCCC}PriViLege & 2024 & CVPR & ViT-B/16 & 86.57M & - & All & VL-Prototype & \xmark & Loss & \checkmark & 95.27 & 1.41 & - & 88.08 & 2.80 & - & 77.50 & 7.13 & - \\
\cellcolor[HTML]{CCCCCC}FineFMPL & 2024 & IJCAI & ViT-B/16 & 86.57M & - & All & VL-Prototype & \xmark & Loss & \checkmark & 93.46 & 2.54 & - & 84.24 & 5.66 & - & 79.75 & 10.30 & - \\
\cellcolor[HTML]{CCCCCC}YourSelf & 2024 & ECCV & ViT-B/16 & 10.0M & 11.90M & All & Prototype & \xmark & Tr. Net. & \xmark & 68.79 & 15.21 & - & 67.02 & 15.88 & - & 69.77 & 19.80 & - \\
\cellcolor[HTML]{CCCCCC}ApproxFSCIL & 2024 & ECCV & ViT-B/16 & 86.57M & - & All & Prototype & \xmark & Loss & \xmark & 91.15 & 2.00 & - & 84.07 & 5.74 & - & 82.67 & 5.39 & - \\ \hline
\end{tabular}
\vspace{-5pt}
\caption{\small \textcolor{black}{Property and Performance of FSCIL methods. Color red, orange, yellow, purple and gray represent backbone tuning approach, meta-learning approach, prototype tuning approach, dynamic structure approach, and PEFT approach respectively, + indicates the substitution of prototype-based classifiers to original classifiers in the respective methods, \^ { } indicates the measurement is computed in the last task.}}
\vspace{-7pt}
\label{tab:main_comparison}
\end{table*}

\begin{wraptable}{r}{0.215\textwidth}
\vspace{-13pt}
\setlength{\tabcolsep}{0.22em}
\centering
\tiny
\begin{tabular}{lcccc}
\hline
\multicolumn{1}{l}{}                                  & \multicolumn{2}{c}{CIFAR100} & \multicolumn{2}{c}{CUB} \\ \cline{2-5}
\multicolumn{1}{l}{\multirow{-2}{*}{\textbf{Method}}} & Base          & Novel        & Base       & Novel      \\ \hline
CEC                                                                             & 67.75         & 21.08        & 70.46      & 34.16      \\
FACT                                                                            & 71.73         & 21.95        & 72.84      & 39.93      \\
C-FSCIL                                                                         & 72.87         & 14.00        & -          & -          \\
LIMIT                                                                           & 71.32         & 20.13        & 73.46      & 42.15      \\
BiDist                                                                          & 64.50         & 20.35        & 70.36      & 39.83      \\
NC-FSCIL                                                                        & 73.98         & 29.30        & 76.19      & 43.07      \\
TEEN                                                                            & 83.68         & 23.47        & 78.67      & 40.81      \\
OrCo                                                                            & 58.22         & 43.15        & 66.62      & 49.25      \\ \hline
L2P                                                                             & 92.07         & 0.00         & 88.00      & 1.67       \\
DualP                                                                           & 91.38         & 0.00         & 89.67      & 3.00       \\
CodaP                                                                           & 93.79         & 0.00         & 90.67      & 3.33       \\
L2P+                                                                            & 82.07         & 57.93        & 80.67      & 67.33      \\
DualP+                                                                          & 82.41         & 68.28        & 81.33      & 72.33      \\
CodaP+                                                                          & 83.45         & 63.45        & 78.33      & 68.00      \\
ASP                                                                             & 91.03         & 80.34        & 86.67      & 80.67     \\ \hline
\end{tabular}
\vspace{-8pt}
\caption{\small \textcolor{black}{Base vs novel classes accuracy in the last task}}
\label{tab:stab_plas_balance}
\vspace{-15pt}
\end{wraptable}
\subsection{Stability-Plasticity Balance}
\vspace{-2pt}
The stability-plasticity balance of FSCIL methods is shown in average harmonic mean accuracy (AHM) in tab.\ref{tab:main_comparison}. The table shows that stability-plasticity is still an open problem in FSCIL, as almost all the methods achieve less than 80\% AHM, and less than 60\% AHM for non-PEFT methods. In addition, the huge different i.e. up to 35\% between AA and AHM of those methods indicates the failure in maintaining stability-plasticity, and the average accuracy relies too much on base classes' accuracy. It indicates that the methods barely achieve plasticity with few available samples. Tab. \ref{tab:stab_plas_balance} shows the accuracy of base and novel classes accuracy in the last task. The table clearly shows the huge (up to 60\%) difference between base and novel classes' accuracy, indicating that the methods rely too much on base class accuracy. PEFT methods with prototype-based classifiers i.e. L2P+ to ASP  achieve a far better balance, shower still suffer from a significant AA-to-AHM gap i.e. up to 25\%.
\vspace{-2pt}
\subsection{Application}
\vspace{-2pt}
Table \ref{tab:fscil_application} summarizes the latest advancement of FSCIL studies in various areas i.e. medical(red), remote sensing (green), NLP (yellow), and graph (blue). Medical and remote sensing typically utilize image data type, while NLP and graph FSCIL typically utilize audio and graph data types respectively. The other data types are time-series, hyperspectral, and text that are utilized in medical, remote sensing, and NLP FSCIL respectively. The majority of the studies were conducted in supervised FSCIL, only few studies were conducted in different settings i.e. UFSCIL\cite{UNISA_10639442}, CFSCIL\cite{GRSMDP_tai2023mine},\cite{NLPAUDIOEDE_si2024fully},  and CDFSCIL\cite{MEDCDFSCIL_yang2023few}. At the moment, backbone tuning, prototype tuning, and dynamic networks are preferable in these areas among the 5 FSCIL approaches, with ResNet as the preferable backbone for all the research areas, except graph FSCIL that utilizes graph neural networks (GNN). The other backbone types are CRF, CNN, and 1DCNN utilized for text, audio, and times series data respectively.       

Prototype-based method is preferred over network-based classifiers due to its resilience against data scarcity and its compatibility to constrastive learning and self-supervised learning. About half of the methods utilize augmentation, but almost none of the methods deploy prototype rectification. It emphasizes the importance of rectification as discussed in the previous analysis. Only of the study utilizes PTM as none of the methods uses PEFT approach where PTM is an important element. None of the studies utilize language-guided learning that shows an opportunity for the approach to be deployed.       

\begin{table}[]
\setlength{\tabcolsep}{0.05em}
\setlength{\belowcaptionskip}{-15pt plus 0pt minus 0pt}
\centering
\tiny
\begin{tabular}{lcccccccc}
\hline
\textbf{Method} & \textbf{Data} & \textbf{Setting} & \textbf{Approach} & \textbf{Backbone} & \textbf{Ptm} & \textbf{Cls.} & \textbf{Aug.} & \textbf{Rect.} \\ \hline
\cellcolor[HTML]{FFA6A6}FS3DCIOT\cite{MEDFS3DCIOT_10226417} & image & FSCIL & Prototype Tun. & ResNet & \xmark & Ptp. & \xmark & Loss \\
\cellcolor[HTML]{FFA6A6}CDFSCIL\cite{MEDCDFSCIL_yang2023few} & image & CDFSCIL & Prototype Tun. & ResNet & \xmark & Ptp. & img. & Loss \\
\cellcolor[HTML]{FFA6A6}MAPIC\cite{MEDMAPIC_10050007} & t.series & FSCIL & Prototype Tun. & 1DCNN & \xmark & Ptp. & feat. & TF. Cal. \\
\cellcolor[HTML]{FFA6A6}FSCIRDR\cite{MEDFSCIRDR_zhang2024few} & image & FSCIL & Prototype Tun. & ResNet & \checkmark & Ptp. & img. & Loss \\ \hline
\cellcolor[HTML]{AFD095}CUC\cite{GRSCUC_zhao2022few} & hyper. & FSCIL & Dynamic Net. & ResNet & \xmark & Net. & \xmark & \xmark \\
\cellcolor[HTML]{AFD095}MDP\cite{GRSMDP_tai2023mine} & image & CFSCIL & Prototype Tun. & ResNet & \xmark & Ptp. & \xmark & Tr. Net. \\
\cellcolor[HTML]{AFD095}UNISA\cite{UNISA_10639442} & image & UFSCIL & Backbone Tun. & ResNet & \xmark & Ptp. & feat. & Loss \\

\cellcolor[HTML]{AFD095}G-MFCN\cite{GRSG-MFCN_wang2024gradient} & image & FSCIL & Prototype Tun. & ResNet & \xmark & Ptp. & feat. & Pse. Ptp. \\
\cellcolor[HTML]{AFD095}SARGRS\cite{GRSSAR_zhao2023few} & image & FSCIL & Prototype Tun. & ResNet & \xmark & Ptp. & feat. & Pse. Ptp. \\ \hline
\cellcolor[HTML]{FFE994}DFSL\cite{NLPAUDIODFSL_wang2021few} & audio & FSCIL & Backbone Tun. & CNN & \xmark & Net. & \xmark & \xmark \\
\cellcolor[HTML]{FFE994}NERFSCIL\cite{NLPNER_wang2022few} & text & FSCIL & Backbone Tun. & CRF & \xmark & Net. & text & \xmark \\
\cellcolor[HTML]{FFE994}SAMP\cite{NLPAUDIOSAMP_li2023few} & audio & FSCIL & Dynamic Net. & ResNet & \xmark & Ptp. & \xmark & Tr. Net. \\
\cellcolor[HTML]{FFE994}EDE\cite{NLPAUDIOEDE_si2024fully} & audio & CFSCIL & Dynamic Net. & ResNet & \xmark & Ptp. & \xmark & Tr. Net. \\
\cellcolor[HTML]{FFE994}AMFO\cite{NLPAUDIOAMFO_li2024few} & audio & FSCIL & Dynamic Net. & ResNet & \xmark & Ptp. & \xmark & Tr. Net. \\ \hline
\cellcolor[HTML]{B4C7DC}GFSCIL\cite{GRAPH_tan2022graph} & graph & FSCIL & Prototype Tun. & GNN & \xmark & Ptp. & \xmark & Tr. Net. \\
\cellcolor[HTML]{B4C7DC}Geometer\cite{GRAPHGEOMETER_lu2022geometer} & graph & FSCIL & Prototype Tun. & GNN & \xmark & Ptp. & \xmark & Tr. Net. \\
\cellcolor[HTML]{B4C7DC}Mecoin\cite{MecoinGRAPHEMM_li2024an} & graph & FSCIL & Prototype Tun. & GNN & \xmark & Ptp. & \xmark & Tr. Net. \\
\cellcolor[HTML]{B4C7DC}Inductive\cite{GRAPHINDUCTIVE_li2024inductive} & graph & FSCIL & Prototype Tun. & GNN & \xmark & Ptp. & tplg & Pse. Ptp. \\ \hline
\end{tabular}
\vspace{-7pt}
\caption{\small \textcolor{black}{Application of FSCIL in various areas. Color red, green yellow, and blue represent medical, remote sensing, NLP and graph research areas respectively.}}
\vspace{-11pt}
\label{tab:fscil_application}
\end{table}

\vspace{-5pt}
\section{Open Challenges and Potential Solutions}
\vspace{-3pt}
\noindent \textbf{a). Stability-Plasticity Dilemma } remains the most important open challenge in FSCIL. The PEFT approach may achieve far better stability-plasticity balance than the other 4 approaches, but the gap between base and novel classes' performance is still high. One of the potential solutions is improving prototype rectification for the PEFT approach since most of its methods utilize rectification by loss only.  

\noindent \textbf{b). Prototype Bias} has a bigger part in the stability-plasticity dilemma as in point a, as the existing models struggle to achieve plasticity. It shows the needs for more effective prototype rectification for FSCIL methods. 

\noindent \textbf{c). Imbalance learning} is one of the overlooked aspects in FSCIl since the current FSCIL settings assume that the base classes are equipped with balanced samples and/or labels. Thus the existing methods haven't been proven yet against class imbalance. Note that class imbalance is a common occurrence in real-world practice e.g. in medical applications, where the normal class has far more samples than particular disease classes. One of the solutions is the use of augmentation thus all base classes have the same training samples or features. The other options are leveraging generalization ability of pre-trained models and adopting CFSCIL method that treats the base task in n-way k-shot manner.   

\noindent \textbf{d). Interpretable Performance Metrics} is one of FSCIL open challenges since almost all the accuracy of FSCIL methods relies on base classes' accuracy. Harmonic mean accuracy gives a more insightful measurement but still can't interpret the gap between base and novel classes' performance. It also doesn't interpret the cause of the measured performance. Backward Transfer (BWT) and Forward Transfer (FWT) \cite{NEWCLMETRICS_diaz2018don} offer more interpretable metrics that explain the contribution of each task to overall performance. Alternatively, BWT and FWT can be measured both in base classes only and novel classes only to obtain more detailed explanation on the impact of each task to those performance categories. The other alternatives are Generalied Accuracy (GAcc) and Area Under Curve (AUC) which are already utilized in \cite{YOURSELF_tang2024rethinking}.      

\noindent \textbf{e). Data Openness and Rehearsal Free Constraint} are important challenges in FSCIL. Many existing methods save novel class exemplars and/or prototypes and then replay them in the upcoming tasks. This could be a breach of data openness policy where data is only accessible (open) in a particular time frame. This issue could be handled by a clear constraint in the problem formulation and experimental setting.   

\noindent \textbf{f). Setting Fairness} is a subtle issue that is rarely discussed in existing FSCIL studies. FSCIL simulations compare several methods with different backbones, trainable parameters size, modality, and other resources that impact the methods' performance raising questions about the fairness aspects of the simulation. Thus it is important to evaluate the competitive methods in comparable resources aside from the official (released) configurations of the methods.

\vspace{-5pt}
\section{Future Direction}
\vspace{-3pt}
\noindent \textbf{a). Toward Data Privacy:} Data privacy now attracts a concern on FSCIL following real-world scenarios where the learning process is conducted by collaborative distributed clients instead of one standalone agent. The agents hold only a partial of the data and class labels. The agents are allowed to exchange their locally trained models but not the data due to data privacy constraint. In a such scenario, FSCIL training is executed in a federated way, and lead to new problem named federated FSCIL (FFSCIL)\cite{FFSCIL_2025federated}.

\noindent \textbf{b). Toward Data Streams:} As in CIL\cite{OCILMYOPIA_xinrui2024forgetting}, learning in data streams is bound to be a concern in FSCIL. The learning agent can't be assumed to always have storage to save the samples and labels, and the learning process will be executed in an online way. This highly possible setting lead to new problem named online FSCIL (OFSCIL) that will be even more challenging than online CIL.

\noindent \textbf{c). Toward Open World Setting:} Nowadays, open-world setting have been a new focus on CIL studies\cite{OWCL_kim2025open} and eventually will be applied in FSCIL in the future. The out-of-distribution (OOD) detection capability is an important feature for a continually learning agent. The agent should be able to differentiate between learned knowledge and not yet learned knowledge. This leads to new sub-study of FSCIL named open-world FSCIL (OWFSCIL).

\noindent \textbf{d). Towards Foundation Models and PEFT:} The advancement of foundation models impacts the FSCIL study significantly as explained in the previous analysis. There is still a room for improvement in its exploration e.g. which pre-trained model is suitable for which dataset, how the pre-trained model contributes in the cross-domain setting, and the development of foundation model for particular domains such as medical images, audio, and graph. Meanwhile, there are many opportunities to develop new PEFT methods such as new promoting structure and learning, or PEFT based on Low-Rank Adaptation (LORA) as in \cite{INFLORA_liang2024inflora}. 

\noindent \textbf{e). Toward Language-Guided Learning:} The recent study\cite{LEAPGEN_ma'sum2025vision} highlights new advancements of language-guided learning in CIL that are likely useful for FSCIL, i.e. the synergy of class-wise and task-wise language prototypes, mixed-up language-and-vision prototype, generative language prototype by LLM decoder, or language as prompt generator offering future opportunity in FSCIL study. 

\vspace{-5pt}
\section{Conclusion}
\vspace{-3pt}
In this study, we propose a comprehensive survey on the latest advancement of the FSCIL topic. We define formal attributes of the setting, a richer yet clear FSCIL topology, and formal objectives of FSCIL approaches. We emphasize the importance of prototype rectification and present its topology. We analyze the broader aspects of FSCIL methods, from their attributes to the detailed performance. We highlight remarkable FSCIL open challenges and potential solutions, and the future direction of FSCIL towards various aspects that open wide opportunities for future study.  
\small
\bibliographystyle{named}
\bibliography{ijcai25}

\end{document}